\documentclass[runningheads]{llncs}

\usepackage{eccv}

\usepackage{eccvabbrv}

\usepackage{graphicx}
\usepackage{booktabs}
\usepackage{multirow}
\usepackage{makecell} 
\usepackage{color}
\usepackage[font=small,labelfont=bf]{caption}

\usepackage[accsupp]{axessibility}  
\usepackage{soul}

\usepackage{hyperref}

\usepackage{orcidlink}
\usepackage{hyperref}

\usepackage{nicematrix} 
\usepackage{multirow} 
\usepackage{booktabs} 
\usepackage{amsmath} 
\usepackage{diagbox} 
\usepackage{array}
\usepackage{float}

\usepackage{arydshln}

\newcommand{\thickhline}{%
    \noalign {\ifnum 0=`}\fi \hrule height 1pt
    \futurelet \reserved@a \@xhline
}

\usepackage{amssymb}
\usepackage{pifont}
\newcommand{\cmark}{\ding{51}}%
\newcommand{\xmark}{\ding{55}}%

\begin{document}

\title{Towards Real-world Event-guided Low-light Video Enhancement and Deblurring}

\titlerunning{Event-guided Low-light Video Enhancement and Deblurring}


\author{Taewoo Kim\orcidlink{0000-0002-8608-9514} \and
Jaeseok Jeong\orcidlink{0000-0002-9836-2979} \and
Hoonhee Cho\orcidlink{0000-0003-0896-6793} \and
Yuhwan Jeong\orcidlink{0009-0002-0279-146X} 
\and
Kuk-Jin Yoon\orcidlink{0000-0002-1634-2756}}

\authorrunning{Kim et al.}

\institute{Korea Advanced Institute of Science and Technology\\
\email{\{intelpro, jason.jeong, gnsgnsgml, jeongyh98, kjyoon\}@kaist.ac.kr}\\
}

\maketitle

\begin{abstract}
In low-light conditions, capturing videos with frame-based cameras often requires long exposure times, resulting in motion blur and reduced visibility. While frame-based motion deblurring and low-light enhancement have been studied, they still pose significant challenges. Event cameras have emerged as a promising solution for improving image quality in low-light environments and addressing motion blur. They provide two key advantages: capturing scene details well even in low light due to their high dynamic range, and effectively capturing motion information during long exposures due to their high temporal resolution. Despite efforts to tackle low-light enhancement and motion deblurring using event cameras separately, previous work has not addressed both simultaneously. To explore the joint task, we first establish real-world datasets for event-guided low-light enhancement and deblurring using a hybrid camera system based on beam splitters. Subsequently, we introduce an end-to-end framework to effectively handle these tasks. Our framework incorporates a module to efficiently leverage temporal information from events and frames. Furthermore, we propose a module to utilize cross-modal feature information to employ a low-pass filter for noise suppression while enhancing the main structural information. Our proposed method significantly outperforms existing approaches in addressing the joint task. Our project pages are available at \url{https://github.com/intelpro/ELEDNet}.
\end{abstract}

\section{Introduction}
\label{sec:intro}
When capturing videos in low-light conditions, the low levels of environmental illumination causes reduced visibility. Typically, under such low-light conditions, cameras resort to using a longer exposure time to compensate for the lack of brightness in the image. Due to the long exposure time, captured images may exhibit undesired blurring artifacts caused by dynamic objects or abrupt camera motion. As such, images taken under low-light environments commonly show both reduced visibility due to diminished illumination and blurring artifacts from dynamic motion simultaneously. Accordingly, it is imperative to jointly address the problem involving both blurring and low-light illumination effects.
\begin{figure*}[!t]
    \centering
    \includegraphics[width=0.85\linewidth]{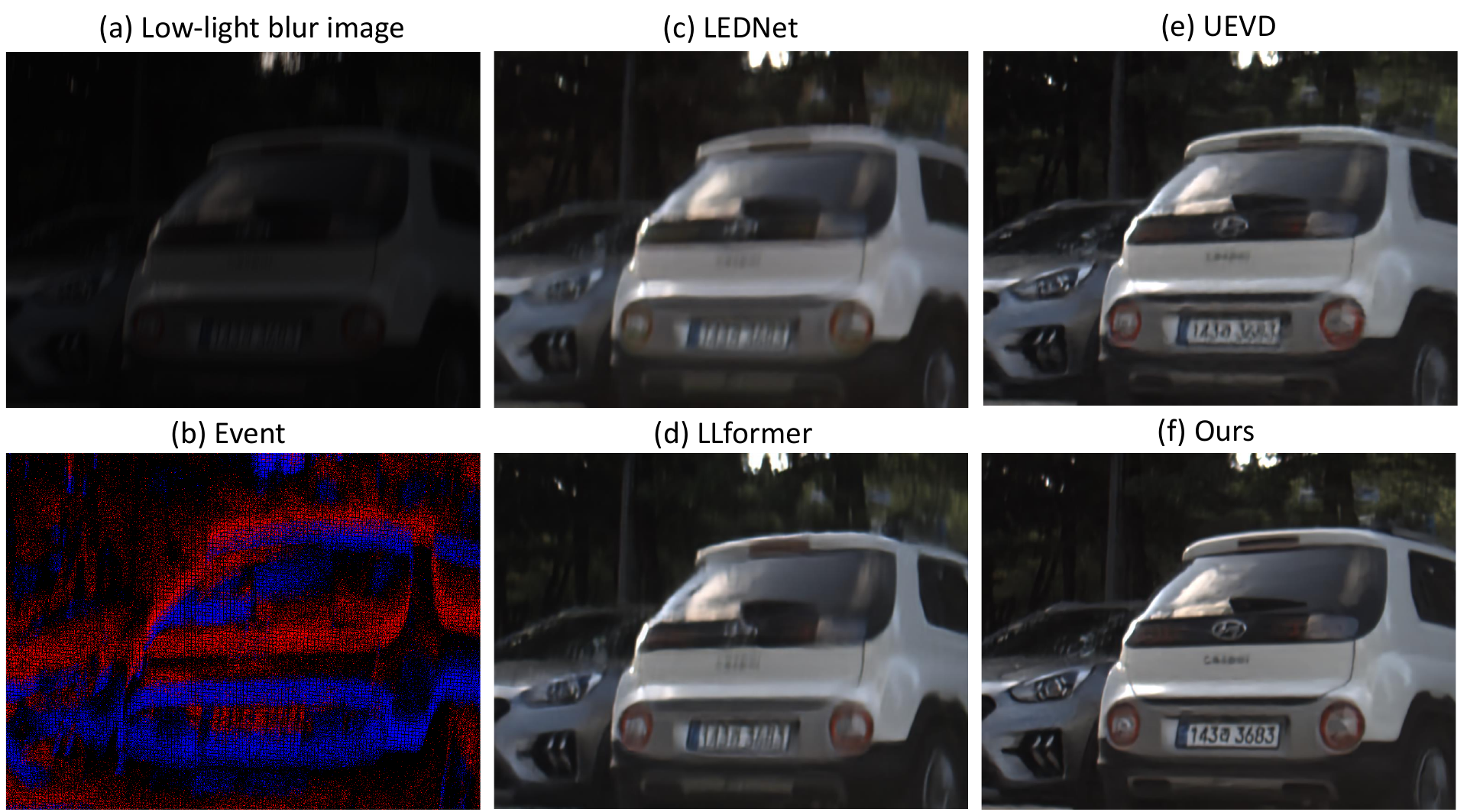}
    \caption{Qualitative comparison with SoTA methods on the real-world low-light blurred images.(Best viewed when zoom in.)
    In figures (c) through (f), visual results are shown for (c) the recent image-based low-light image enhancement and deblurring networks LEDNet~\cite{zhou2022lednet}; (d) the recent low-light enhancement network LLFormer~\cite{wang2023llformer}; (e) event-guided video deblurring networks UEVD~\cite{kim2022event_UEVD}; and (f)Ours, respectively.}
    \label{fig:teaser}
    \centering
\end{figure*}

Numerous endeavors have explored low-light enhancement~\cite{xu2022snr,retinexformer,wang2021sdsd,liu2023low}, aimed at restoring low-light images to a more natural appearance akin to those captured under normal lighting conditions, and motion deblurring~\cite{Zamir2021MPRNet,Zamir2021Restormer,cho2021rethinking,pan2023deep,kim2022event_UEVD}
, focused on eliminating motion blur artifacts from blurred images, 
as distinct and independent tasks. Consequently, these works propose solutions specifically tailored for their respective tasks. 
Though the two tasks, motion deblurring and low-light enhancement, can be performed in a cascaded manner, conducting correlated tasks in a such manner often yields sub-optimal results. As a result, it is essential to address the problem in a joint manner that considers both the occurrence of motion blur and the low-illumination scenario simultaneously rather than dealing with them as isolated issues. To address these limitations, we need to first tackle the joint problem of low-light enhancement and motion deblurring in a unified single network. This task becomes extremely challenging when relying solely on frame information, particularly as the intensity of motion blur increases or in severe low-light conditions where visibility is significantly reduced. In such scenarios, it becomes difficult to restore a normal light sharp image.

The event camera~\cite{gallego2020event_survey}, renowned for its high temporal resolution and unique functionality, has spurred significant research in low-level vision. Its asynchronous capture of luminance changes offers benefits such as high dynamic range, low latency, and low power consumption. The high temporal resolution events capture detailed motion information, aiding in motion deblurring and revealing structural details even in low-light conditions.

Since event cameras excel in event-guided motion deblurring and low-light enhancement individually, they can be offered as practical solutions for jointly addressing both challenges. Hence, we propose a novel task aiming to solve two tasks simultaneously in low-illumination environments utilizing events. However, acquiring real-world datasets for this task is highly challenging. Though recent works have introduced synthetic dataset~\cite{zhou2022lednet} for low-light scenarios, these methods struggle to restore real-motion blur, limiting their generalizability to real world scenarios. To address these limitations, we build the Real-world Event-guided Low-light Video Enhancement and Deblurring(RELED) dataset, which provides real-world synchronized low-light blur, normal-light sharp images, and low-light event streams, without relying on synthetic data generation methods.

In addition to the dataset contribution part, we propose a unified end-to-end framework for event-guided low-light video enhancement and deblurring. We introduce an Event-guided Deformable Temporal Feature Alignment(ED-TFA) module guided by events. While temporal alignment tasks have been applied in various low-level vision tasks, performing temporal alignment solely using multiple video frames in degraded conditions poses a highly ill-posed problem. Due to low visibility and blur, finding temporal correspondence among frames becomes challenging. However, event data retains high dynamic range and temporal resolution, aiding in finding temporal correspondence. Hence, we propose the ED-TFA module, effectively utilizing event information to incorporate temporal information across multiple visual scales.

Secondly, we propose a novel Spectral Filtering-based Cross-Modal Feature Enhancement(SFCM-FE) module. Several studies in event-guided low-level vision~\cite{sun2022event,sun2023event_REFID,zhang2023generalizing} have explored cross-modal feature enhancement. However, in low-illumination situations, both events and blurred images are affected by noise, posing challenges in restoring main structural details.
To more effectively capture main structures from cross-modality features, we introduce the SFCM-FE module, which utilizes low-frequency information. This module aims to enhance the extraction of structural details from low-frequency features while simultaneously reducing noise induced by low-light conditions.

To summarize, our contributions are four-folds: 
(1) Our works marks the first attempt to simultaneously address event-guided low-light enhancement and deblurring problem.
(II) To address this new challenge, we developed a novel hybrid-camera system and collected a distinctive real-world dataset: the RELED dataset.
(III) Finally, we introduce an end-to-end framework designed to effectively tackle the joint task. Specifically, our framework consists of two modules: the Event-guided Deformable Temporal Feature Alignment (ED-TFA) module and the Spectral Filtering-based Cross-Modal Feature Enhancement (SFCM-FE) module.
(IV) We achieve state-of-the-art performance when compared to other networks, showcasing the effectiveness of our frameworks in this novel task.
\section{Related works}
\subsection{Frame-based and Event-guided Motion Deblurring}
From the advancements in deep learning, there have been significant developments in motion deblurring research. Early research(\eg,~\cite{sun2015learning}) focused on estimating the blur kernel to derive a non-blind motion deblurring model. After the emergence of datasets for motion deblurring(\eg, GoPro~\cite{nah2017deep}), research on various network architectures for motion deblurring gained momentum. For example, ~\cite{cho2021rethinking,nah2017deep} proposed architectures based on multi-scale to increase the receptive field of CNNs. Subsequently, ~\cite{zhang2019deep,Zamir2021MPRNet,tao2018scale} utilized network architectures based on multi-patch to enhance performance. Following this, thanks to advancements in the transformer architecture~\cite{vaswani2017attention}, research on image deblurring utilizing transformer-based approaches such as restormer~\cite{Zamir2021Restormer} and stripformer~\cite{tsai2022stripformer} has been proposed. Recently, some works~\cite{kong2023efficient,dong2023multi,mao2023intriguing} have also been introduced in deblurring that leverage the advantages in the frequency domain.

However, motion deblurring remains a challenging task due to the degradation of motion information caused by motion blur, rendering it highly ill-posed. Despite this challenge, event cameras offer a promising solution with their low-latency feature, enabling them to capture dense motion information within the exposure time of frame-based cameras. Pan \etal.~\cite{pan2019bringing_edi_cvpr,pan2020high_edi} proposed the event-based double integral model (EDI) for event-guided motion deblurring, but it often yields suboptimal results due to event camera noise. Subsequent studies aimed to improve event-guided deblurring, with approaches including recurrent networks~\cite{jiang2020learning}, physical model-based networks~\cite{lin2020_LEDVDI}, and cross-modal attention~\cite{sun2022event,sun2023event_REFID}, unknown exposure time videos~\cite{kim2022event_UEVD}, non-coaxial settings~\cite{cho2023non}, spike streams~\cite{chen2024enhancing}. \textit{Unlike these works, we aim to address joint low-light enhancement and motion deblurring using event cameras.}
\subsection{Frame-based and Event-guided Low-light Enhancement}
Since the emergence of deep learning, significant progress has been made in Low Light Enhancements(LLE)~\cite{li2021low_survey}. Initially, Lore~\etal~\cite{lore2017llnet} proposed a method using a sparse denoising auto-encoder to convert low-light images to normal light ones.  Subsequently, Wei \etal~\cite{wei2018deep} introduced the LOL dataset, the first dataset comprising pairs of normal-light and low-light images.  As real-world low-light enhancement datasets were lacking, supervised learning became feasible for training low-light enhancement methods. Wang \etal.\cite{wang2019underexposed} enhanced low-light images by estimating intermediate illumination maps instead of using direct end-to-end network mapping. 
Wu~\etal\cite{wu2022uretinex} improved efficiency by employing deep unfolding networks based on the retinex theory. Xu~\etal~\cite{xu2022snr} proposed a network architecture that simultaneously utilizes transformer and CNN structures by incorporating the Signal-to-Noise Ratio(SNR). Recently, there has been a surge in research on LLE, exploring techniques such as gamma correction~\cite{wang2023low}, retinex-based transformers~\cite{retinexformer}, implicit neural representations~\cite{yang2023inr_lle}, diffusion models~\cite{yi2023diff,wang2023exposurediffusion}, among others~\cite{liu2023low,zheng2023empowering}.

Recent research has explored low-light enhancement using event cameras, leveraging their high temporal resolution and dynamic range. Liu \etal~\cite{liu2023low_synthetic} proposed a framework that generates synthetic events from multiple frames and utilizes event information for low-light enhancement. Liang~\etal.\cite{liang2023coherent} further extended this by incorporating event information alongside frames through multimodal coherence modeling and temporal coherence propagation. 

\begin{figure*}[!t]
    \centering
    \includegraphics[width=1.0\linewidth]{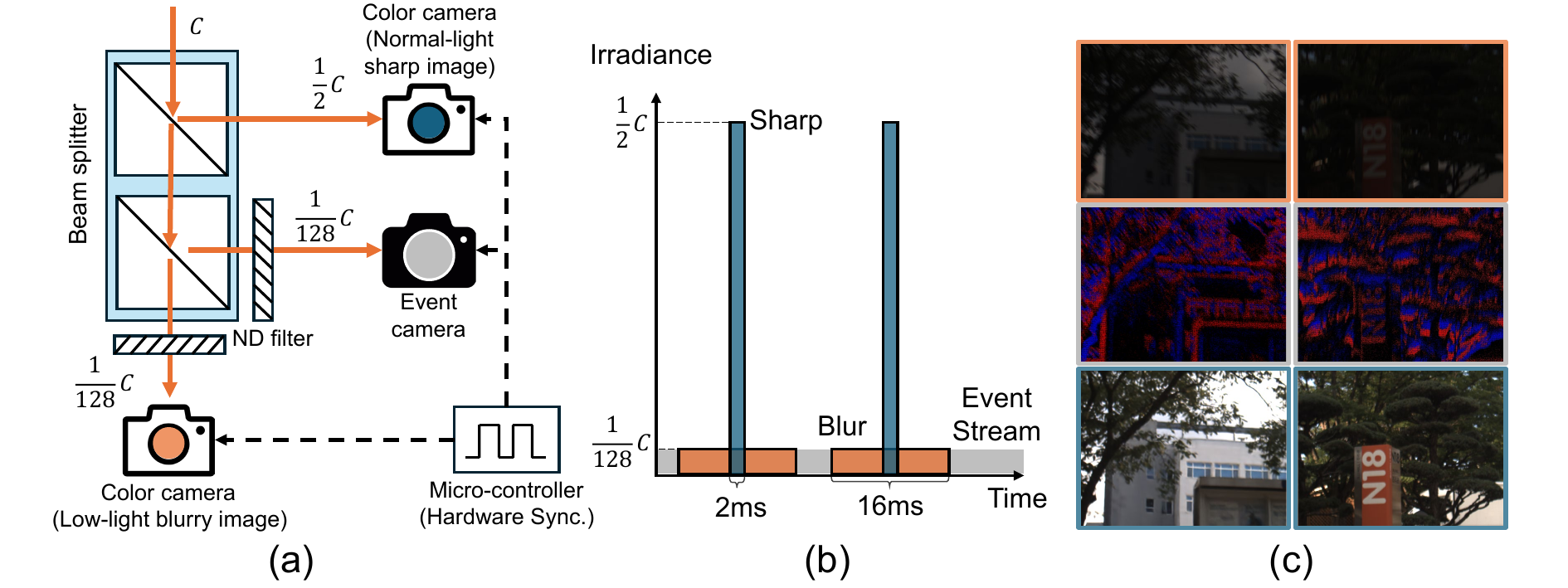}
    \caption{In the Figure, (a),(b) and (c) respectively represents the schematic diagram of our hybrid camera system. the exposure time scheme of color cameras and samples of our datasets.
    In (c), from top to bottom, each represents low-light blurry image, event stream, and normal light sharp image, respectively.
    }
    \label{fig:camera_config}
\end{figure*}

\section{RELED Dataset}
\subsection{Limitation of Other LLE and Deblurring datasets.}
\noindent
\textbf{Low-light Enhancement(LLE)} 
Previous LLE methods~\cite{lore2017llnet, lv2018mbllen} use gamma correction to simulate low-light images from normal-light ones, but these differ from real low-light images. Network-based methods (\eg~\cite{guo2020zero}) aim to bridge this gap. Recent approaches use the HDR advantage of event cameras for event-guided LLE, simulating both low-light images and corresponding events. However, methods~\cite{liang2023coherent,liu2023low_synthetic} relying on synthetic generation~\cite{rebecq2018esim,hu2021v2e} struggle to model real-world low-light scenarios.
\\
\noindent
\textbf{Motion Deblurring}
Researchers often use synthesized blur images from high frame rate videos like GoPro~\cite{nah2017deep} or DVD datasets~\cite{su2017deep} to train and evaluate frame-based motion deblurring methods. These images are averages of consecutive frames and do not account for photometric effects like saturation, noise, and dynamic ranges. Event cameras, with their high temporal resolution, are also used in motion deblurring. Unlike synthetic methods, the REBlur dataset~\cite{sun2022event} captures real-world blur but is limited in capturing dynamic scenes and suffers from low resolution. To improve event-guided motion deblurring, higher resolution and real-world datasets are needed.
In summary, the main challenges with existing event-guided motion deblurring and the LLE dataset are the lack of real-world datasets, including real events, and the low resolution of cameras, such as DAVIS cameras.

\subsection{RELED dataset acquisition}
While some datasets have been proposed for event-based low-level vision~\cite{sun2022event,sun2023event_REFID}, there have been no attempts to jointly address event-guided LLE and motion deblurring. To perform this new event-guided low-light video enhancement and deblurring task, we need to simultaneously acquire synchronized low-light blur images, normal-light sharp images, and the corresponding event stream. Several studies have proposed various datasets using different camera systems~\cite{rim_2020_ECCV,zhong2020efficient,zhong2021towards} to capture real world datasets . However, obtaining pairs of low-light blurry images and normal sharp images, along with synchronized event streams simultaneously, remains a challenging problem.
Prior works~\cite{zhou2022lednet} utilize data synthesis methods, such as ZeroDCE~\cite{guo2020zero} or gamma correction, to model low-light settings, however, there still exist limitations when generalizing to real-world low-light conditions. 

\begin{table}[!t]
\caption{Comparison of RELED datasets with publicly available datasets.~\cite{zhou2022lednet}}
\resizebox{\columnwidth}{!}{
\setlength{\tabcolsep}{3pt}
\begin{tabular}{c|ccccc}
\hline
Datasets & Image resolution & color & Real event(resolution) & Low-light type & Blur type  \\ \hline
LoL-Blur~\cite{zhou2022lednet} & 1120$\times$640 & Yes & Not provide & Synthetic low-light & Synthetic blur  \\
RELED(Ours) & \textbf{1024$\times$768} &  \textbf{Yes} & \textbf{Yes(1024$\times$768)} & \textbf{Real low-light} & \textbf{Real blur}  \\ \hline
\end{tabular}
}
\label{tab:RELED_info}
\end{table}
To this end, we build the RELED (\textbf{R}eal-world \textbf{E}vent-guided \textbf{L}ow-light video \textbf{E}nhancement and \textbf{D}eblurring) dataset without relying on synthetic generation of low-light images and events. To simultaneously capture normal-light images, low-light images, and the corresponding event stream, we designed a triple-axis beam-splitter-based camera system, as shown in Fig.~\ref{fig:camera_config}. This system comprises two high-resolution RGB cameras and one event camera. One RGB camera captures sharp images under normal-light conditions, while the other captures blurred images in low-light conditions. In our camera setup, each beam splitter splits the incoming light in half, with the camera attached to the first beam splitter receiving half of the original intensity, and the cameras attached to the second beam splitter receiving one-fourth of the original intensity.

To make the best use of our configuration, the color camera at the first beam splitter captures normal-light sharp images employing a short exposure time(2ms),  while the color camera at the second beam splitter captures blurred images through long exposure times(16ms). Though the intensity at the second camera is half the intensity of the first camera, the longer exposure time causes the second camera to capture more light rays. As such, additional ND filters are added to limit the light intensity to induce low-light conditions. With a 1/32 ND filter, the resulting incident intensity at the event camera and the second color camera is 1/128 of the original intensity. Taking into consideration the exposure time, the beam splitter, and the ND filters, the radiant exposure at the second color camera is 1/8 the radiant exposure at the first camera. 

We achieve hardware-level synchronization of multiple devices using a micro-controller. The beam splitter aligns all cameras mechanically, but minor misalignments require additional calibration via a homography matrix. This results in the RELED dataset, featuring low-light blur images, normal-light sharp images, and low-light event streams. We captured 42 urban scenes, including camera movement and moving objects. The dataset, sized at $1024 \times 768$, is the first to offer high-resolution images with real-world low-light blur and normal-light sharp images, and we believe it will be valuable for the research community.


\begin{figure*}[!t]
    \centering
    \includegraphics[width=0.9\linewidth]{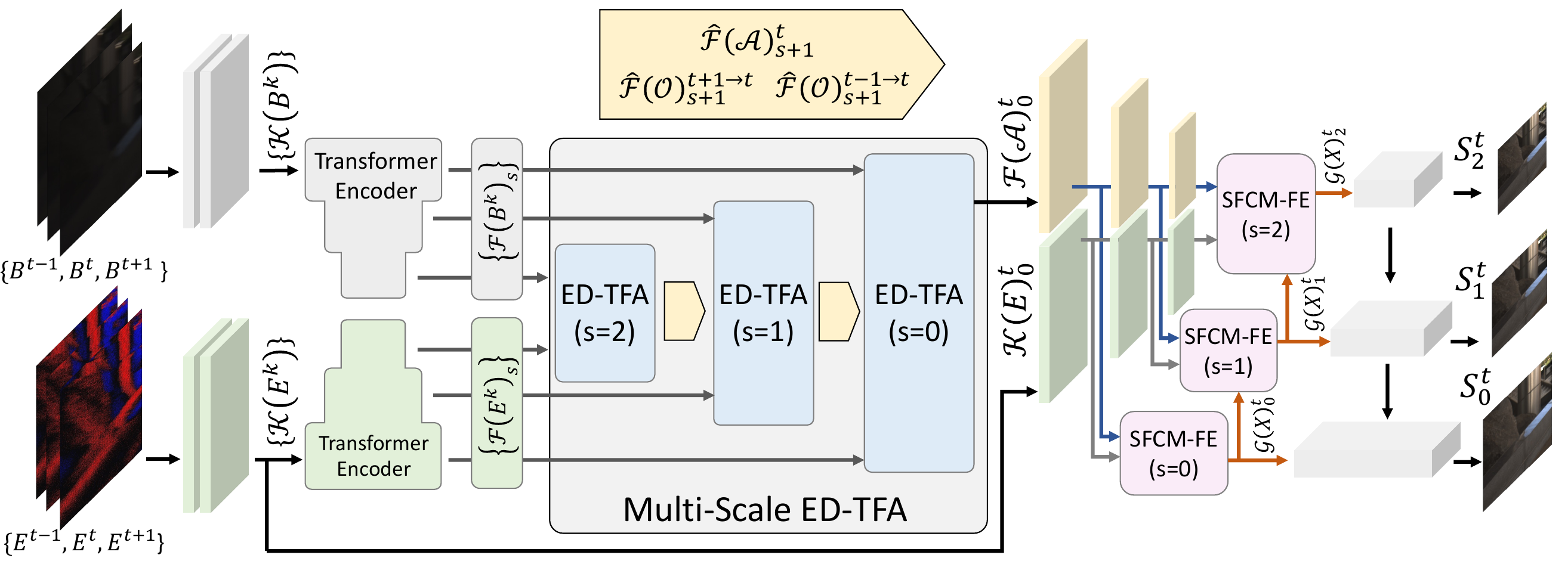}
    \caption{Overall framework of the proposed methods. In the figure, the subscript numbers below each feature represent scale factor, while the superscript indicates the timestamp index.}
    \label{fig:overall_frameworks}
\end{figure*}

\section{Proposed Methods}
\subsection{Overall framework}
The overall framework of the proposed method is illustrated in Fig.~\ref{fig:overall_frameworks}. 
We convert the event stream into voxel grid representation~\cite{Zhu_2019_CVPR} due to its superior spatio-temporal preservation ability. We represent event voxel grid as $\mathrm{E}^{k} \in \mathbb{R}^{B \times H \times W}$ at timestamps index $k$ with voxel bin size $B$. Our networks utilize three sequential blur frames $\{B^{t-1},B^{t},B^{t+1}\}$ and sequential event voxels $\{E^{t-1},E^{t}, E^{t+1}\}$.
Utilizing these two modality inputs, we first extract temporal event features $\{\mathcal{K}(E)^{k}\}$ and blur features $\{\mathcal{K}(B)^{k}\}$ for each timestamp $k$ through several convolution layer where $k \in \{t-1, t, t+1\}$.
Our proposed framework consists of two major modules in total.

The first module, Event-guided Deformable Temporal Feature Alignment (ED-TFA) module generates temporally aligned features. To this end, we first extract features from blur features $\{\mathcal{K}(B^{k})\}$ and event features $\{\mathcal{K}(E^{k})\}$ using a transposed attention~\cite{Zamir2021MPRNet}-based transformer encoder. This produces blur feature pyramid $\{\mathcal{F}(B^{k})_{s}\}$ and event feature pyramid $\{\mathcal{F}(E^{k})_{s}\}$. These features from the two modality feature pyramids are passed into ED-TFA of various scales, with intermediary upscaled offset information sequentially passed between ED-TFA modules of different scales. Finally, the ED-TFA module at the original scale ($s=0$) generates aligned feature $\mathcal{F(A)}_{s=0}$ with scale factor $s$ of 0.

The second module, the Spectral Filtering-based Cross-Modal Frequency Feature Enhancement(SFCM-FE) module, involves encoding the aligned feature $\mathcal{F}(A)^{t}_{s=0}$ and $\mathcal{K}(E)^{t}_{s=0}$, which yields the event feature pyramid $\{\mathcal{G}(E)^{t}_{s}\}$ and aligned feature pyramid $\{\mathcal{G}(A)^{t}_{s}\}$ through a CNN-based encoder~\cite{Zamir2021MPRNet}. The goal of the SFCM-FE module is to efficiently utilize the cross-modality feature pyramid by capitalizing on the benefits of the spectral filtering process.
It generates $\mathcal{G}(X)^{t}_{s}$ using the aligned feature $\mathcal{G}(A)^{t}_{s}$ and event feature $\mathcal{G}(E)^{t}_{s}$ and the upsampled output of SFCM-FE module estimated at previous scale $\hat{\mathcal{G}}(X)^{t}_{s+1}$.
After the SFCM-FE module, the generated enhanced cross-modality feature pyramids, denoted as $\{\mathcal{G}(X)_{s}^{t}\}$, are then passed to the UNet-based decoder, ultimately resulting in the generation of multi-scale outputs $\{S^{t}_{s}\}$ where $s \in \{0,1,2\}$.

\begin{figure*}[!t]
    \centering
    \includegraphics[width=0.9\linewidth]{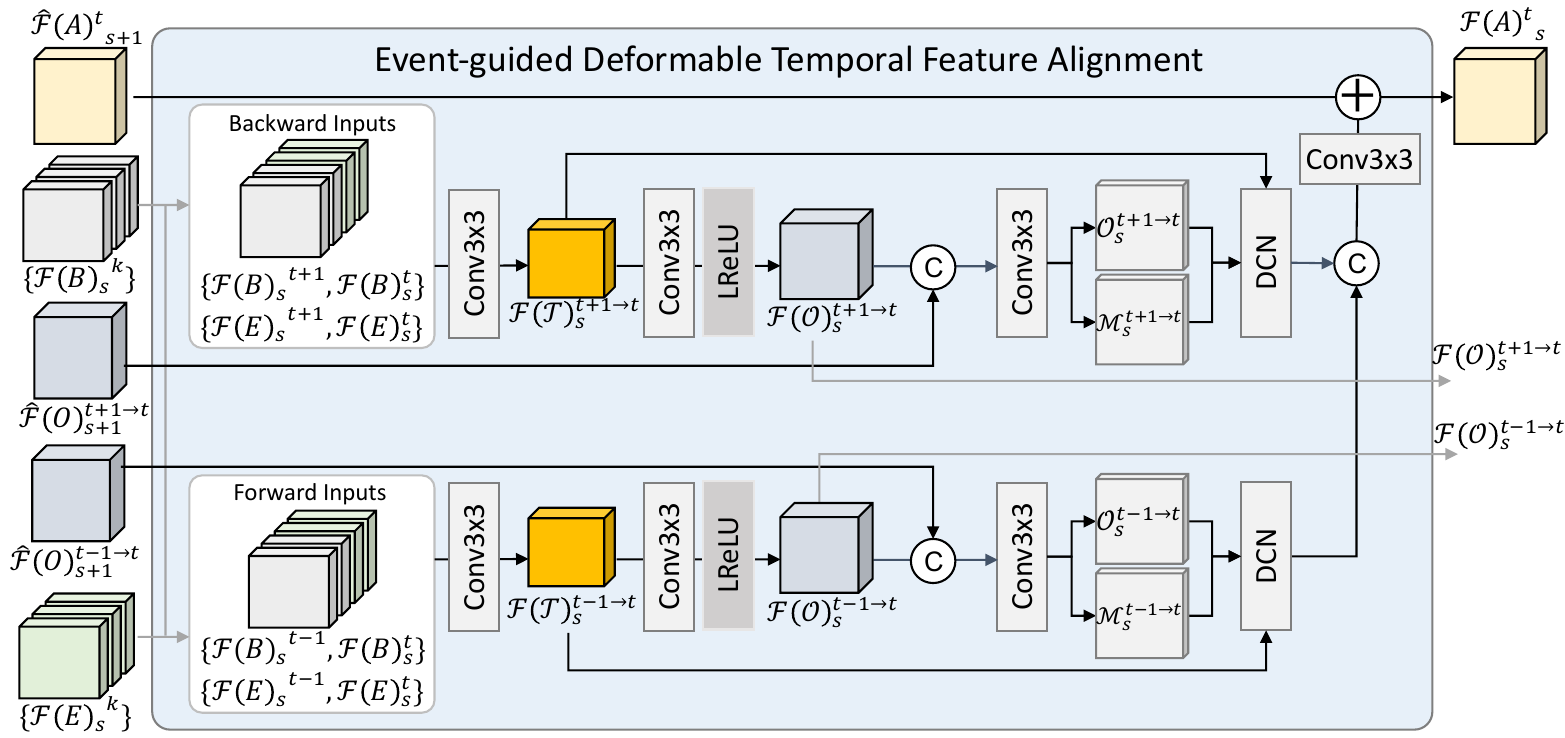}
    \caption{The overall structure of proposed ED-TFA module.}
    \label{fig:ED_TFA}
\end{figure*}

\subsection{Event-guided Deformable Temporal Feature Alignment(ED-TFA) Module.}
Temporal alignment is the process of extracting valuable information from multiple video frames. Alignment methods have been proven effective in fields such as video deblurring~\cite{zhong2020efficient,chao2022rnnmbp,zhou2019spatio,pan2023deep} and LLE~\cite{wang2021sdsd} works. In situations where motion blur and low-light conditions coexist, as in our task, temporal alignment becomes even more challenging: the significant amount of noise in the frames as well as motion blur make finding dense correspondence between frames extremely difficult. 

In low-light conditions, events can capture dense motion information with high temporal resolution, thanks to their high dynamic range characteristics, greatly aiding in temporal feature alignment. While several event-guided video deblurring methods~\cite{kim2022event_UEVD,sun2023event_REFID} have employed multiple video frames, these methods typically involve concatenating multiple video frames for the encoding phase. 
However, concatenating multiple video frames often leads to sub-optimal temporal alignment. To address this, we propose the ED-TFA module, which uses event guidance for deformable temporal alignment. 
Initially, event and frame features are extracted using a transposed attention~\cite{Zamir2021Restormer}-based transformer encoder known for better capturing non-local context information and computational efficiency at high resolution.
The features are aligned in a coarse-to-fine manner across various scales of ED-TFA module.
At lower resolutions, sub-pixel misalignment is minimal, but it increases with higher resolutions. The multi-scale ED-TFA modules continuously transmit intermediary information to make ongoing adjustments, leading to more robust alignment.
While we conduct both forward and backward alignment, as depicted in Fig.~\ref{fig:ED_TFA}, we explain only one as both steps are identical and only differ in input.

To perform backward alignment on the $t$ frame, we first synthesize template features used as input for alignment. These features are generated using both the future frame feature, $\mathcal{F}(B)_{s}^{t+1}$, and the current frame features $\mathcal{F}(B)_{s}^{t}$, along with the current and future event features $\mathcal{F}(E)_{s}^{t}$ and $\mathcal{F}(E)_{s}^{t+1}$, respectively.
Finally, all concatenated features pass through the convolution layer to generate the template feature $\mathcal{F(T)}^{t+1 \rightarrow t}_{s}$.
For deformable alignment, we first estimate the feature $\mathcal{F(O)}_{s}^{t+1 \rightarrow t}$ to determine the offset and modulation masks. 
Then, we utilize the previously estimated offset feature information $\mathcal{F(O)}_{s+1}^{t+1 \rightarrow t}$ by applying transposed convolution to upsample the feature, resulting in $\mathcal{\hat{F}(O)}_{s+1}^{t+1 \rightarrow t}$; here the hat$(\hat{\mathcal{F}})$ symbol represents features that have been upsampled using transposed convolution.
Subsequently, we concatenate these two features and estimate the modulation masks $\mathcal{M}_{s}^{t+1\rightarrow t}$ and offsets $\mathcal{O}_{s}^{t+1\rightarrow t}$ for deformable convolution (DCN)~\cite{zhu2019deformable}. The application of DCN $\mathcal{D}$ is as follows:

\begin{equation}
\mathcal{F(T)}_{s}^{t+1 \rightarrow t} = \mathcal{D}(\mathcal{F(T)}_{s}^{t+1 \rightarrow t}, \mathcal{O}_{s}^{t+1 \rightarrow t}, \mathcal{M}_{s}^{t+1 \rightarrow t})
\end{equation}
where $\mathcal{D}$ denotes a deformable convolution.
In a similar fashion, we apply deformable alignment to obtain forward aligned features $\mathcal{F(T)}_{s}^{t-1 \rightarrow t}$ as in the Fig.\ref{fig:ED_TFA}. 
Ultimately, we estimate the residual of the upsampled temporally aligned feature $\mathcal{\hat{F}(A)}_{s}^{t}$ by concatenating forward-aligned features $\mathcal{F(T)}^{t-1 \rightarrow t}_{s}$ and backward-aligned features $\mathcal{F(T)}^{t+1 \rightarrow t}$, and applying a convolutional layer. The aligned feature $\mathcal{F(A)}^{t}_{s}$, with the backward offset feature $\mathcal{F(O)}_{s}^{t+1 \rightarrow t}$ and forward offset feature $\mathcal{F(O)}_{s}^{t-1 \rightarrow t}$, serve as inputs for the ED-TFA module as shown in Fig.\ref{fig:overall_frameworks}.

\begin{figure*}[!t]
    \centering
    \includegraphics[width=0.85\linewidth]{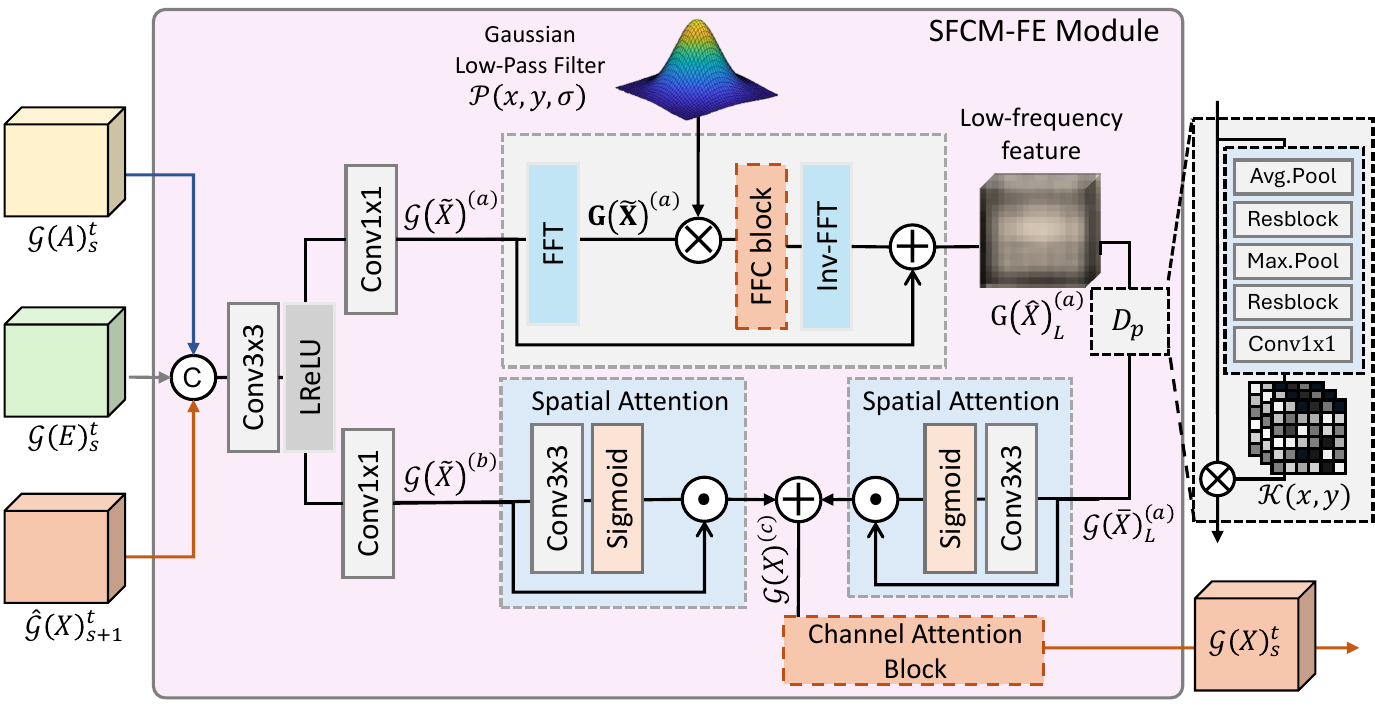}
    \caption{The overall structure of proposed SFCM-FE module.}
    \label{fig:FFCM_FE}
\end{figure*}
\subsection{Spectral Filtering-based Cross-Modal Feature Enhancement(SFCM-FE) Module.}
Several event-guided low-level vision works~\cite{sun2022event, sun2023event_REFID, chen2024enhancing} have explored cross-modality feature fusion to leverage the complementary strengths of event and image modalities. Images typically provide detailed semantic information and structural insights, while event data offer high temporal resolution motion information and rich high-frequency components of images. To effectively exploit the complementary advantages of event and image information, various approaches have been proposed for cross-modality feature fusion.

However, integrating these feature fusion methods into our network may not lead to performance improvement in scenarios where motion blur and low-illumination conditions coexist, as in our task. This is because, in low-illumination conditions where motion blur is present, a low signal-to-noise ratio (SNR) results in significant noise generation in the blurred images, making it extremely challenging to restore global structural information. Additionally, events, with their abundant high dynamic range properties, effectively capture scene details in low-illumination situations. However, as the illumination level of the scene decreases, events also tend to generate a significant amount of noise. Therefore, it is necessary to develop a cross-modality feature enhancement method capable of effectively reducing such high levels of noise and accurately restoring the main structural information of the scene. To address these challenges, we propose a novel Spectral Filtering-based Cross-Modal Feature Enhancement (SFCM-FE) module. As depicted in Fig.~\ref{fig:FFCM_FE}, we leverage the advantages of a coarse-to-fine approach by first performing upsampling on the feature processed by the SFCM-FE module at the previous scale ($s+1$).
\begin{equation}
\hat{\mathcal{G}}(X)^{t}_{s+1} = \mathrm{DConv}_{4\times4}(\mathcal{G}(X)^{t}_{s+1})
\end{equation}
where $\mathcal{G}(X)^{t}_{s+1}$ denotes the enhanced feature at the previous scale ($s+1$), and $\mathrm{Dconv}_{4\times4}$ denotes a $4\times4$ deconvolution layer, respectively.

After this step, it generates the upsampled feature $\hat{\mathcal{G}}(X)^{t}_{s}$. Subsequently, the event feature $\mathcal{G}(E)^{t}_{s}$, the aligned frame feature $\mathcal{G}(A)^{t}_{s}$, and the upsampled feature $\mathcal{G}(X)^{t}_{s+1}$ are concatenated and passed through a convolution layer and a LeakyReLU layer to generate $\mathcal{G}(\tilde{X})$ (omitting temporal index $t$ and scale factor $s$ for brevity). Now that the feature $\mathcal{G}(\tilde{X})$ containing all information has been generated, we pass it through convolution layers to divide it into two branches: One feature $\mathcal{G}(\tilde{X})^{(a)}$ is used for enhancing low-frequency information and suppressing noise through a low-pass filter branch, 
and the other feature $\mathcal{G}(\tilde{X})^{(b)}$ passes through a different branch. To pass through the low-frequency branch, we perform frequency domain filtering using a low-pass filter. First, we apply the Fast Fourier Transforms (FFT) to the feature $\mathcal{G}(\tilde{X})^{(a)}$, denoted as $\mathbf{G(\tilde{X})^{(a)}} = \mathrm{FFT}(\mathcal{G}(\tilde{X})^{(a)})$. Next, we apply a Gaussian low-pass filter~\cite{liu2022global} to extract low-frequency information. The Gaussian filter is defined as follows:

\begin{equation}
\mathcal{P}(x,y,\sigma) = \mathrm{exp}(-\frac{(x-x_{c})^{2}+(y-y_c)^{2}}{2\sigma^{2}})
\end{equation}
where $x_{c}$ and $y_{c}$ represent the center point in the spectral domain along the x and y axes, respectively, and $\sigma$ represents the standard deviation.
The gaussian filter, with a value of 1 at the center and decreasing values as it moves away from the center, can be applied as a low-pass filter. 
We apply low-pass filter $\mathcal{P}(x,y,\sigma)$ to the feature $\mathbf{G(\tilde{X})}^{(a)}$ as follows:
\begin{equation}
\mathbf{G}\mathbf{(\hat{X})}_{L}^{(a)} = \mathcal{P}(x,y,\sigma) \odot \mathbf{G(\tilde{X}})^{(a)}
\end{equation}
where $\odot$ denote element-wise multiplication. 
After low-pass filtering, we pass through an FFC~\cite{chi2020fast}(Fast Fourier Convolution) block to perform additional frequency selection in the spectral domain($\mathrm{Conv_{1x1}}+\mathrm{ReLU}+\mathrm{Conv_{1x1}}$).
Finally, we apply inverse-FFT to inverse the transformation, making the frequency-filtered feature.  We then produce $\mathrm{G}(\hat{X})^{(a)}_{L}$ through a residual connection with the original feature $\mathrm{G}(X)^{(a)}$.  The features $\mathrm{G}(\hat{X})^{(a)}_{L}$ emphasized by low-frequency information typically correspond to spatially variant main structural information. To better enhance spatially variant main structure of the scenes, we apply pixel-wise spatial dynamic filters~\cite{zhou2021decoupled,pan2023deep} to the feature $\mathrm{G}_{L}(\hat{X})^{(a)}$. To achieve this, we utilize the pixel-wise dynamic filter block $\mathcal{F}_{s}$ to estimate pixel-wise spatial dynamic filter $\mathcal{K}(x,y)$ with filter size ($f_{k} \times f_{k}$).
\begin{equation}
\mathcal{G}_{L}(\bar{X})^{(a)} = \mathcal{K}(x,y)\otimes \mathrm{G}_{L}(\hat{X})^{(a)}
\end{equation}
where $\otimes$ and $\mathcal{G}_{L}(\bar{X})^{(a)}$ denote dynamic convolution operation and output filtered feature, respectively. Through this step, we have generated features capable of enhancing the structural information of low frequencies and removing noise information associated with low illumination. Finally, we perform feature fusion between the information containing the original features without frequency filtering and the residual learning-based information. 
During feature fusion, we apply spatial attention~\cite{woo2018cbam} to each feature to remove irrelevant spatial information as follows:
\begin{equation}
\mathcal{G}(X)^{(c)} = \mathcal{G}(\bar{X})_{L}^{(a)} \odot \mathcal{\sigma}(\mathrm{Conv}_{3\times3}(\mathcal{G}(\bar{X})_{L}^{(a)})) + \mathcal{G}(\tilde{X})^{(b)}\odot \mathcal{\sigma}(\mathrm{Conv}_{3\times3}(\mathcal{G}(\tilde{X})^{(b)}))
\label{eq:spatial_attention}
\end{equation}
where $\mathcal{\sigma}$ denotes sigmoid function.
Lastly, $\mathcal{G}(X)^{(c)}$ is fed into  multiple Channel Attention Blocks(CAB)~\cite{Zamir2021MPRNet}, resulting in the generation of the enhanced feature representation, denoted as $\mathcal{G}(X)_{s}$.  
This feature, which has passed through the SFCM-FE module, is used as the input for the next scale SFCM-FE module and simultaneously serves as input for the decoder, as shown in Fig.~\ref{fig:overall_frameworks}.

\begin{figure*}[t]
    \centering
    \includegraphics[width=1.0\linewidth]{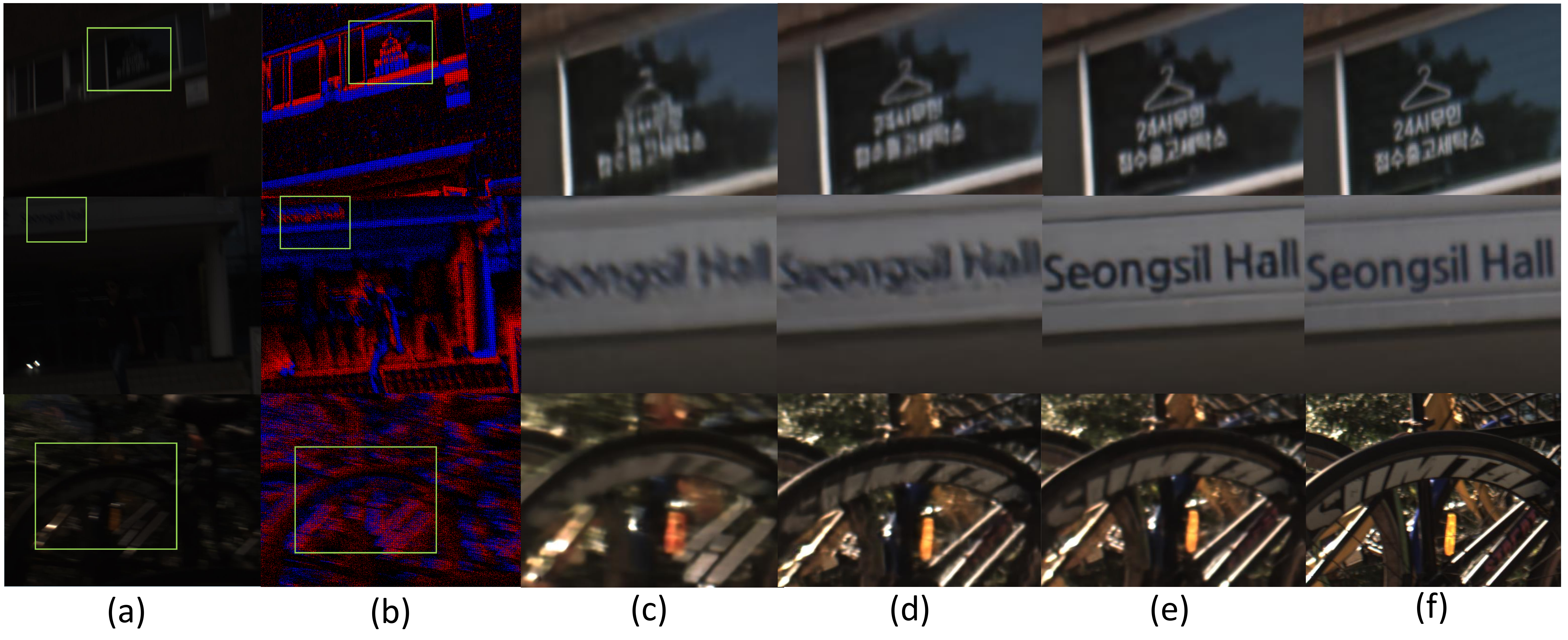}
    \caption{Visual comparisons on the RELED datasets. In the figures, (a) to (f) depict the following: (a) input low-light blurry images, (b) low-light events, (c) LLFormer~\cite{wang2023llformer}, (d) EFNet~\cite{sun2022event}, (e) Ours, and (f) GT normal-light sharp image. }
    \label{fig:visual_results}
\end{figure*}

\section{Experiments}
\subsection{Datasets and baseline methods}
Since there was no dataset available containing synchronized events along with normal light sharp ground truth and low-light blur inputs, we conducted all experiments using the RELED dataset. The RELED dataset comprises 29 training scenes and 13 test scenes, providing low-light blurred input paired directly with normal light sharp ground truth.
Consequently, we trained all baseline methods end-to-end. For benchmarking against the open-source LLE network, we utilized state-of-the-art image-based methods such as Retinexformer~\cite{retinexformer}, LLFormer~\cite{wang2023llformer}, SNRNet~\cite{xu2022snr}, and the video-based method SDSDNet~\cite{wang2021sdsd}. 
For motion deblurring networks, we utilized image-based deblurring networks including MPRNet~\cite{Zamir2021MPRNet}, MIMONet+~\cite{cho2021rethinking}, and NAFNet~\cite{chen2022simple}, along with video-based methods RNN-MBP~\cite{chao2022rnnmbp} and DSTNet~\cite{pan2023deep}. 
As a joint method, we conducted comparisons with the only existing work, LEDNet~\cite{zhou2022lednet}. Finally, we compared with event-guided image-deblurring methods such as e-SLNet~\cite{wang2020event}, REDNet~\cite{xu2021motion_REDNet}, EFNet~\cite{sun2022event}, and event-guided video deblurring methods such as REFID~\cite{sun2023event_REFID} and UEVD~\cite{kim2022event_UEVD}.

\subsection{Experimental results on the RELED Datasets}
For a fair comparison, all networks were trained and tested using the splits of the RELED dataset, with training conducted for a total of 200 epochs. The quantitative comparison results on the RELED dataset are presented in Tab.~\ref{tab:real_quatitative} with the table divided into frame-based LLE methods, frame-based motion deblurring methods, frame-based joint methods, and event-based deblurring methods. In conditions where both motion blur and low illumination are prevalent simultaneously, frame-based methods did not achieve satisfactory performance. Compared to frame-based LLE methods, we observed a PSNR performance gap ranging from 2.83 dB to 4.83 dB. Similarly, compared to frame-based deblurring methods, we achieved significant performance improvements, with PSNR improvements ranging from 1.17 dB to 4.78 dB. Additionally, compared to the joint LLE and deblurring method, LEDNet~\cite{zhou2022lednet}, we observed a significant gain of 4.83 dB. While event-guided methods generally outperform frame-based methods due to advantages of events to capture scene structures effectively even in low-light conditions and provide high-temporal resolution motion information, our method still achieved performance improvements ranging from 11.85 dB to 1.20 dB in terms of PSNR compared to event-guided methods. This improvement can be attributed to our task-oriented network design.
Moreover, our lightweight model, ours-s, outperformed other networks while using only a relatively small number of parameters (5.3MB).
Qualitative comparisons depicted in Fig.~\ref{fig:teaser} and ~\ref{fig:visual_results} consistently demonstrate superior performance of our network, even in challenging scenarios such as low-light conditions with significant noise in both frames and events, as well as severe motion blur situations.

\begin{table}[!t]
\centering
\caption{Quantitative evaluation on the RELED dataset. $\dagger$ denotes the event-guided method. The \textbf{Bold} and \ul{underline} denote the best and the second-best performance.}
\label{tab:real_quatitative}
\scalebox{0.85}{
\resizebox{\columnwidth}{!}{
\setlength{\tabcolsep}{18pt}
\begin{tabular}{c|cc|cc}
\hline
Methods & PSNR & SSIM & Params(MB)  \\ \hline
SNRNet~\cite{xu2022snr} & 26.47 &  0.851 & 40.08 \\
SDSDNet~\cite{wang2021sdsd} & 28.47 & 0.887 & 4.43  \\
LLFormer~\cite{wang2023llformer} & 26.62 & 0.862 &  13.15 \\
RetinexFormer~\cite{retinexformer} & 26.66 & 0.865 & 1.61 \\ \hline
MPRNet~\cite{Zamir2021MPRNet} & 26.89 & 0.867 & 20.13  \\
MIMOUNet+~\cite{cho2021rethinking} & 26.52 & 0.866 & 16.11  \\
NAFNet~\cite{chen2022simple} & 26.77 & 0.862 & 67.91  \\
RNN-MBP~\cite{chao2022rnnmbp} & 29.52 & 0.902  & 14.16  \\
DSTNet~\cite{pan2023deep} & 29.59 & 0.903 & 7.53 \\ \hline
LEDNet~\cite{zhou2022lednet} & 26.47 & 0.856 & 7.41 \\ \hline
e-SLNet$^{\dagger}$~\cite{wang2020event} & 19.45 & 0.663 & 0.17 \\
REDNet$^{\dagger}$~\cite{xu2021motion_REDNet} & 29.19 & 0.903 & 9.7  \\
EFNet$^{\dagger}$~\cite{sun2022event} & 29.85 & 0.905 & 8.47 \\
UEVD$^{\dagger}$~\cite{kim2022event_UEVD} & 29.93 & 0.905 & 27.88 \\
GEM$^{\dagger}$~\cite{zhang2023generalizing} & 26.04 & 0.810 & 2.36 \\
REFID$^{\dagger}$~\cite{sun2023event_REFID} & 30.10 & 0.913 & 15.9 \\ \hline
Ours-s$^{\dagger}$ & \underline{30.98} & \underline{0.919} & 5.3 \\ 
Ours$^{\dagger}$ & \textbf{31.30} & \textbf{0.925} & 12.8 \\ 
\hline
\end{tabular}
}
}
\centering
\end{table}

\subsection{Model Analysis}
To analyze the network components of our method, we conducted an ablation study for each module. All ablation experiments were conducted on the RELED dataset, training different versions of each network for a total of 200 epochs. We reported the ablation results of each module in Tab.~\ref{tab:main_ablations}. In the baseline model, blur feature $\mathcal{F}(B)^{t}_{s=0}$ is directly passed instead of $\mathcal{F}(\mathcal{A})^{t}_{s=0}$ to our UNet architecture while SFCM-FE modules are replaced with Conv 1$\times$1 layers.

\noindent
\textbf{ED-TFA module.} 
We compared the performance of the baseline network with and without the ED-TFA module. This comparison showed a significant performance gain of +1.19 dB in terms of PSNR. Furthermore, when comparing the network version (Ver.3) with the SFCM-FE module inserted into the baseline to the full model (Ver.4), we observed a performance improvement of +0.90 dB. These results highlight the significance of our ED-TFA module. \\
\noindent
\textbf{SFCM-FE module.}
We evaluated the impact of the SFCM-FE module on performance by comparing the baseline network (Ver.1) with network (Ver.3), which integrates the SFCM-FE module. This analysis showed a significant performance improvement of +0.81 dB. Additionally, when comparing network Ver.2 and Ver.4, both with the ED-TFA module, we observed a significant performance improvement of +0.52 dB. \\
\noindent\textbf{In-depth analysis of the SFCM-FE module.}
In the ablation study of the SFCM-FE module, reported in Tab.\ref{Tab:SFCM_FE}, 
inclusion of either CABs~\cite{Zamir2021MPRNet} or SA (Eq.~\ref{eq:spatial_attention}) showed negligible performance improvement (+0.03dB/+0.01dB).
However, the addition of network blocks for low-frequency modulation and spatial attention (Ver.4) to evaluate the performance enhancement of Low-Pass Filter(LPF) branch resulted in a significant improvement of +0.44 dB. Further enhancement was observed when including CAB block (Ver.5), resulting in an additional improvement of +0.08 dB compared to Ver.4.

\begin{table}[!t]
\centering
\caption{Ablation study of the network components on the RELED datasets}
\scalebox{0.9}{
\setlength{\tabcolsep}{5pt}
\begin{tabular}{c|cccc|cc}
\hline
Methods & Baseline & ED-TFA & SFCM-FE &  PSNRs & \#Params(MB) \\ \hline
Ver.1 & \cmark & \xmark & \xmark &  29.59 & 1.8\\
Ver.2 & \cmark & \cmark & \xmark & 30.78\textcolor{red}{\scriptsize{(+1.19$\uparrow$)}} &  5.0 \\
Ver.3 & \cmark & \xmark & \cmark &  30.40\textcolor{red}{\scriptsize{(+0.81$\uparrow$)}} & 9.7 \\
Ver.4 & \cmark & \cmark & \cmark &  31.30\textcolor{red}{\scriptsize{(+1.71$\uparrow$)}} &  12.8 \\ 
\hline
\end{tabular}
}
\centering
\label{tab:main_ablations}
\end{table}

\begin{table}[!t]
\centering
\caption{An ablation study was conducted on the components of the SFCM-FE module. The LPF branch, representing the Low Pass Filter branch, corresponds to the feature modulation branch marked with superscript (a) in Fig.~\ref{fig:FFCM_FE}.}
\scalebox{0.9}{
\setlength{\tabcolsep}{15pt}
\begin{tabular}{c|ccc|c}
\hline
Methods & CABs & SA & LPF branch & PSNRs \\ \hline
Ver.1 & \xmark & \xmark & \xmark & 30.78 \\
Ver.2 & \cmark & \xmark & \xmark & 30.81\textcolor{red}{\scriptsize{(+0.03$\uparrow$)}} \\
Ver.3 & \cmark & \cmark & \xmark & 30.79\textcolor{red}{\scriptsize{(+0.01$\uparrow$)}} \\
Ver.4 & \xmark & \cmark & \cmark & 31.22\textcolor{red}{\scriptsize{(+0.44$\uparrow$)}} \\
Ver.5 & \cmark & \cmark & \cmark &  31.30\textcolor{red}{\scriptsize{(+0.52$\uparrow$)}} \\ \hline
\end{tabular}
}
\centering
\label{Tab:SFCM_FE}
\end{table}

\begin{table}[!t]
\centering
\caption{Comparisons on the event-image cross-modality feature fusion module.}
\scalebox{0.9}{
\setlength{\tabcolsep}{3pt}
\begin{tabular}{c|cccc}
\hline
Methods & w/o Fusion & w/ EFNet~\cite{sun2022event} Fusion & w/ REFID~\cite{sun2023event_REFID} Fusion & SFCM-FE(Ours) \\ \hline
PSNRs &  30.78 & 30.55\textcolor{blue}{\scriptsize{(-0.23$\downarrow$)}} & 30.86\textcolor{red}{\scriptsize{(+0.08$\uparrow$)}} & 31.3\textcolor{red}{\scriptsize{(+0.52$\uparrow$)}} \\
\hline
\end{tabular}
}
\centering
\label{Tab:other_fusion}
\end{table}
\noindent\textbf{Comparison of the feature fusion module.}
We evaluated alternative event-image feature fusion modules like EFNet~\cite{sun2022event} and REFID~\cite{sun2023event_REFID} in place of the SFCM-FE module, as shown in Tab.~\ref{Tab:other_fusion}. However, this replacement resulted in minimal performance changes, highlighting our SFCM-FE module.

\section{Conclusion}
This paper addresses the novel research problem of event-guided low-light video enhancement and deblurring. To achieve this, we designed a hybrid camera system using beam splitters and constructed the RELED dataset containing low-light blurry images, normal sharp images, and event streams. Subsequently, we developed a tailored framework for the task and validated its effectiveness. Finally, we achieved significant performance improvement on the proposed dataset, surpassing both event-guided and frame-based methods.

\noindent
\textbf{Acknowledgements.}
This work was supported by the National Research Foundation of Korea(NRF) grant funded by the Korea government(MSIT)(NRF2022R
1A2B5B03002636).

%
%
\bibliographystyle{splncs04}

\begin{thebibliography}{10}
\providecommand{\url}[1]{\texttt{#1}}
\providecommand{\urlprefix}{URL }
\providecommand{\doi}[1]{https://doi.org/#1}

\bibitem{retinexformer}
Cai, Y., Bian, H., Lin, J., Wang, H., Timofte, R., Zhang, Y.: Retinexformer: One-stage retinex-based transformer for low-light image enhancement. In: ICCV (2023)


\bibitem{chao2022rnnmbp}
Chao, Z., Hang, D., Jinshan, P., Boyang, L., Yuhao, H., Lean, F., Fei, W.: Deep recurrent neural network with multi-scale bi-directional propagation for video deblurring. In: AAAI (2022)

\bibitem{chen2022simple}
Chen, L., Chu, X., Zhang, X., Sun, J.: Simple baselines for image restoration. In: European Conference on Computer Vision. pp. 17--33. Springer (2022)

\bibitem{chen2024enhancing}
Chen, S., Zhang, J., Zheng, Y., Huang, T., Yu, Z.: Enhancing motion deblurring in high-speed scenes with spike streams. Advances in Neural Information Processing Systems  \textbf{36} (2024)

\bibitem{chi2020fast}
Chi, L., Jiang, B., Mu, Y.: Fast fourier convolution. Advances in Neural Information Processing Systems  \textbf{33},  4479--4488 (2020)

\bibitem{cho2023non}
Cho, H., Jeong, Y., Kim, T., Yoon, K.J.: Non-coaxial event-guided motion deblurring with spatial alignment. In: Proceedings of the IEEE/CVF International Conference on Computer Vision. pp. 12492--12503 (2023)

\bibitem{cho2021rethinking}
Cho, S.J., Ji, S.W., Hong, J.P., Jung, S.W., Ko, S.J.: Rethinking coarse-to-fine approach in single image deblurring. In: Proceedings of the IEEE/CVF international conference on computer vision. pp. 4641--4650 (2021)

\bibitem{dong2023multi}
Dong, J., Pan, J., Yang, Z., Tang, J.: Multi-scale residual low-pass filter network for image deblurring. In: Proceedings of the IEEE/CVF International Conference on Computer Vision. pp. 12345--12354 (2023)

\bibitem{gallego2020event_survey}
Gallego, G., Delbr{\"u}ck, T., Orchard, G., Bartolozzi, C., Taba, B., Censi, A., Leutenegger, S., Davison, A.J., Conradt, J., Daniilidis, K., et~al.: Event-based vision: A survey. IEEE transactions on pattern analysis and machine intelligence  \textbf{44}(1),  154--180 (2020)

\bibitem{guo2020zero}
Guo, C., Li, C., Guo, J., Loy, C.C., Hou, J., Kwong, S., Cong, R.: Zero-reference deep curve estimation for low-light image enhancement. In: Proceedings of the IEEE/CVF conference on computer vision and pattern recognition. pp. 1780--1789 (2020)

\bibitem{hu2021v2e}
Hu, Y., Liu, S.C., Delbruck, T.: v2e: From video frames to realistic dvs events. In: Proceedings of the IEEE/CVF Conference on Computer Vision and Pattern Recognition. pp. 1312--1321 (2021)

\bibitem{jiang2020learning}
Jiang, Z., Zhang, Y., Zou, D., Ren, J., Lv, J., Liu, Y.: Learning event-based motion deblurring. In: Proceedings of the IEEE/CVF Conference on Computer Vision and Pattern Recognition. pp. 3320--3329 (2020)


\bibitem{kim2022event_UEVD}
Kim, T., Lee, J., Wang, L., Yoon, K.J.: Event-guided deblurring of unknown exposure time videos. In: European Conference on Computer Vision. pp. 519--538. Springer (2022)

\bibitem{kong2023efficient}
Kong, L., Dong, J., Ge, J., Li, M., Pan, J.: Efficient frequency domain-based transformers for high-quality image deblurring. In: Proceedings of the IEEE/CVF Conference on Computer Vision and Pattern Recognition. pp. 5886--5895 (2023)

\bibitem{li2021low_survey}
Li, C., Guo, C., Han, L., Jiang, J., Cheng, M.M., Gu, J., Loy, C.C.: Low-light image and video enhancement using deep learning: A survey. IEEE transactions on pattern analysis and machine intelligence  \textbf{44}(12),  9396--9416 (2021)

\bibitem{liang2023coherent}
Liang, J., Yang, Y., Li, B., Duan, P., Xu, Y., Shi, B.: Coherent event guided low-light video enhancement. In: Proceedings of the IEEE/CVF International Conference on Computer Vision. pp. 10615--10625 (2023)

\bibitem{lin2020_LEDVDI}
Lin, S., Zhang, J., Pan, J., Jiang, Z., Zou, D., Wang, Y., Chen, J., Ren, J.: Learning event-driven video deblurring and interpolation. In: Computer Vision--ECCV 2020: 16th European Conference, Glasgow, UK, August 23--28, 2020, Proceedings, Part VIII 16. pp. 695--710. Springer (2020)

\bibitem{liu2023low_synthetic}
Liu, L., An, J., Liu, J., Yuan, S., Chen, X., Zhou, W., Li, H., Wang, Y.F., Tian, Q.: Low-light video enhancement with synthetic event guidance. In: Proceedings of the AAAI Conference on Artificial Intelligence. vol.~37, pp. 1692--1700 (2023)

\bibitem{liu2022global}
Liu, Y., Yu, R., Wang, J., Zhao, X., Wang, Y., Tang, Y., Yang, Y.: Global spectral filter memory network for video object segmentation. In: European Conference on Computer Vision. pp. 648--665. Springer (2022)

\bibitem{liu2023low}
Liu, Y., Huang, T., Dong, W., Wu, F., Li, X., Shi, G.: Low-light image enhancement with multi-stage residue quantization and brightness-aware attention. In: Proceedings of the IEEE/CVF International Conference on Computer Vision. pp. 12140--12149 (2023)

\bibitem{lore2017llnet}
Lore, K.G., Akintayo, A., Sarkar, S.: Llnet: A deep autoencoder approach to natural low-light image enhancement. Pattern Recognition  \textbf{61},  650--662 (2017)

\bibitem{lv2018mbllen}
Lv, F., Lu, F., Wu, J., Lim, C.: Mbllen: Low-light image/video enhancement using cnns. In: BMVC. vol.~220, p.~4 (2018)

\bibitem{mao2023intriguing}
Mao, X., Liu, Y., Liu, F., Li, Q., Shen, W., Wang, Y.: Intriguing findings of frequency selection for image deblurring. In: Proceedings of the AAAI Conference on Artificial Intelligence. vol.~37, pp. 1905--1913 (2023)

\bibitem{nah2017deep}
Nah, S., Hyun~Kim, T., Mu~Lee, K.: Deep multi-scale convolutional neural network for dynamic scene deblurring. In: Proceedings of the IEEE conference on computer vision and pattern recognition. pp. 3883--3891 (2017)

\bibitem{pan2023deep}
Pan, J., Xu, B., Dong, J., Ge, J., Tang, J.: Deep discriminative spatial and temporal network for efficient video deblurring. In: Proceedings of the IEEE/CVF Conference on Computer Vision and Pattern Recognition. pp. 22191--22200 (2023)

\bibitem{pan2020high_edi}
Pan, L., Hartley, R., Scheerlinck, C., Liu, M., Yu, X., Dai, Y.: High frame rate video reconstruction based on an event camera. IEEE Transactions on Pattern Analysis and Machine Intelligence  (2020)

\bibitem{pan2019bringing_edi_cvpr}
Pan, L., Scheerlinck, C., Yu, X., Hartley, R., Liu, M., Dai, Y.: Bringing a blurry frame alive at high frame-rate with an event camera. In: Proceedings of the IEEE/CVF Conference on Computer Vision and Pattern Recognition. pp. 6820--6829 (2019)

\bibitem{rebecq2018esim}
Rebecq, H., Gehrig, D., Scaramuzza, D.: Esim: an open event camera simulator. In: Conference on robot learning. pp. 969--982. PMLR (2018)

\bibitem{rim_2020_ECCV}
Rim, J., Lee, H., Won, J., Cho, S.: Real-world blur dataset for learning and benchmarking deblurring algorithms. In: Computer Vision--ECCV 2020: 16th European Conference, Glasgow, UK, August 23--28, 2020, Proceedings, Part XXV 16. pp. 184--201. Springer (2020)

\bibitem{su2017deep}
Su, S., Delbracio, M., Wang, J., Sapiro, G., Heidrich, W., Wang, O.: Deep video deblurring for hand-held cameras. In: Proceedings of the IEEE conference on computer vision and pattern recognition. pp. 1279--1288 (2017)

\bibitem{sun2015learning}
Sun, J., Cao, W., Xu, Z., Ponce, J.: Learning a convolutional neural network for non-uniform motion blur removal. In: Proceedings of the IEEE conference on computer vision and pattern recognition. pp. 769--777 (2015)

\bibitem{sun2022event}
Sun, L., Sakaridis, C., Liang, J., Jiang, Q., Yang, K., Sun, P., Ye, Y., Wang, K., Gool, L.V.: Event-based fusion for motion deblurring with cross-modal attention. In: European Conference on Computer Vision. pp. 412--428. Springer (2022)

\bibitem{sun2023event_REFID}
Sun, L., Sakaridis, C., Liang, J., Sun, P., Cao, J., Zhang, K., Jiang, Q., Wang, K., Van~Gool, L.: Event-based frame interpolation with ad-hoc deblurring. In: Proceedings of the IEEE/CVF Conference on Computer Vision and Pattern Recognition. pp. 18043--18052 (2023)

\bibitem{tao2018scale}
Tao, X., Gao, H., Shen, X., Wang, J., Jia, J.: Scale-recurrent network for deep image deblurring. In: Proceedings of the IEEE conference on computer vision and pattern recognition. pp. 8174--8182 (2018)

\bibitem{tsai2022stripformer}
Tsai, F.J., Peng, Y.T., Lin, Y.Y., Tsai, C.C., Lin, C.W.: Stripformer: Strip transformer for fast image deblurring. In: European Conference on Computer Vision. pp. 146--162. Springer (2022)

\bibitem{vaswani2017attention}
Vaswani, A., Shazeer, N., Parmar, N., Uszkoreit, J., Jones, L., Gomez, A.N., Kaiser, {\L}., Polosukhin, I.: Attention is all you need. Advances in neural information processing systems  \textbf{30} (2017)

\bibitem{wang2020event}
Wang, B., He, J., Yu, L., Xia, G.S., Yang, W.: Event enhanced high-quality image recovery. In: European Conference on Computer Vision. Springer (2020)

\bibitem{wang2021sdsd}
Wang, R., Xu, X., Fu, C.W., Lu, J., Yu, B., Jia, J.: Seeing dynamic scene in the dark: A high-quality video dataset with mechatronic alignment. In: Proceedings of the IEEE/CVF international conference on computer vision. pp. 9700--9709 (2021)

\bibitem{wang2019underexposed}
Wang, R., Zhang, Q., Fu, C.W., Shen, X., Zheng, W.S., Jia, J.: Underexposed photo enhancement using deep illumination estimation. In: Proceedings of the IEEE/CVF conference on computer vision and pattern recognition. pp. 6849--6857 (2019)

\bibitem{wang2023llformer}
Wang, T., Zhang, K., Shen, T., Luo, W., Stenger, B., Lu, T.: Ultra-high-definition low-light image enhancement: A benchmark and transformer-based method. In: Proceedings of the AAAI Conference on Artificial Intelligence. vol.~37, pp. 2654--2662 (2023)

\bibitem{wang2023low}
Wang, Y., Liu, Z., Liu, J., Xu, S., Liu, S.: Low-light image enhancement with illumination-aware gamma correction and complete image modelling network. In: Proceedings of the IEEE/CVF International Conference on Computer Vision. pp. 13128--13137 (2023)

\bibitem{wang2023exposurediffusion}
Wang, Y., Yu, Y., Yang, W., Guo, L., Chau, L.P., Kot, A.C., Wen, B.: Exposurediffusion: Learning to expose for low-light image enhancement. In: Proceedings of the IEEE/CVF International Conference on Computer Vision. pp. 12438--12448 (2023)

\bibitem{wei2018deep}
Wei, C., Wang, W., Yang, W., Liu, J.: Deep retinex decomposition for low-light enhancement. arXiv preprint arXiv:1808.04560  (2018)

\bibitem{woo2018cbam}
Woo, S., Park, J., Lee, J.Y., Kweon, I.S.: Cbam: Convolutional block attention module. In: Proceedings of the European conference on computer vision (ECCV). pp. 3--19 (2018)

\bibitem{wu2022uretinex}
Wu, W., Weng, J., Zhang, P., Wang, X., Yang, W., Jiang, J.: Uretinex-net: Retinex-based deep unfolding network for low-light image enhancement. In: Proceedings of the IEEE/CVF conference on computer vision and pattern recognition. pp. 5901--5910 (2022)

\bibitem{xu2021motion_REDNet}
Xu, F., Yu, L., Wang, B., Yang, W., Xia, G.S., Jia, X., Qiao, Z., Liu, J.: Motion deblurring with real events. In: Proceedings of the IEEE/CVF International Conference on Computer Vision. pp. 2583--2592 (2021)

\bibitem{xu2022snr}
Xu, X., Wang, R., Fu, C.W., Jia, J.: Snr-aware low-light image enhancement. In: Proceedings of the IEEE/CVF conference on computer vision and pattern recognition. pp. 17714--17724 (2022)

\bibitem{yang2023inr_lle}
Yang, S., Ding, M., Wu, Y., Li, Z., Zhang, J.: Implicit neural representation for cooperative low-light image enhancement. In: Proceedings of the IEEE/CVF International Conference on Computer Vision. pp. 12918--12927 (2023)

\bibitem{yi2023diff}
Yi, X., Xu, H., Zhang, H., Tang, L., Ma, J.: Diff-retinex: Rethinking low-light image enhancement with a generative diffusion model. In: Proceedings of the IEEE/CVF International Conference on Computer Vision. pp. 12302--12311 (2023)

\bibitem{Zamir2021Restormer}
Zamir, S.W., Arora, A., Khan, S., Hayat, M., Khan, F.S., Yang, M.H.: Restormer: Efficient transformer for high-resolution image restoration. In: CVPR (2022)

\bibitem{Zamir2021MPRNet}
Zamir, S.W., Arora, A., Khan, S., Hayat, M., Khan, F.S., Yang, M.H., Shao, L.: Multi-stage progressive image restoration. In: CVPR (2021)

\bibitem{zhang2019deep}
Zhang, H., Dai, Y., Li, H., Koniusz, P.: Deep stacked hierarchical multi-patch network for image deblurring. In: Proceedings of the IEEE/CVF Conference on Computer Vision and Pattern Recognition. pp. 5978--5986 (2019)

\bibitem{zhang2023generalizing}
Zhang, X., Yu, L., Yang, W., Liu, J., Xia, G.S.: Generalizing event-based motion deblurring in real-world scenarios. In: ICCV (2023)

\bibitem{zheng2023empowering}
Zheng, N., Zhou, M., Dong, Y., Rui, X., Huang, J., Li, C., Zhao, F.: Empowering low-light image enhancer through customized learnable priors. In: Proceedings of the IEEE/CVF International Conference on Computer Vision. pp. 12559--12569 (2023)

\bibitem{zhong2020efficient}
Zhong, Z., Gao, Y., Zheng, Y., Zheng, B.: Efficient spatio-temporal recurrent neural network for video deblurring. In: Computer Vision--ECCV 2020: 16th European Conference, Glasgow, UK, August 23--28, 2020, Proceedings, Part VI 16. pp. 191--207. Springer (2020)

\bibitem{zhong2021towards}
Zhong, Z., Zheng, Y., Sato, I.: Towards rolling shutter correction and deblurring in dynamic scenes. In: Proceedings of the IEEE/CVF Conference on Computer Vision and Pattern Recognition. pp. 9219--9228 (2021)

\bibitem{zhou2021decoupled}
Zhou, J., Jampani, V., Pi, Z., Liu, Q., Yang, M.H.: Decoupled dynamic filter networks. In: Proceedings of the IEEE/CVF Conference on Computer Vision and Pattern Recognition. pp. 6647--6656 (2021)

\bibitem{zhou2022lednet}
Zhou, S., Li, C., Loy, C.C.: Lednet: Joint low-light enhancement and deblurring in the dark. In: ECCV (2022)

\bibitem{zhou2019spatio}
Zhou, S., Zhang, J., Pan, J., Xie, H., Zuo, W., Ren, J.: Spatio-temporal filter adaptive network for video deblurring. In: Proceedings of the IEEE/CVF international conference on computer vision. pp. 2482--2491 (2019)

\bibitem{Zhu_2019_CVPR}
Zhu, A.Z., Yuan, L., Chaney, K., Daniilidis, K.: Unsupervised event-based learning of optical flow, depth, and egomotion. In: Proceedings of the IEEE/CVF Conference on Computer Vision and Pattern Recognition (CVPR) (June 2019)

\bibitem{zhu2019deformable}
Zhu, X., Hu, H., Lin, S., Dai, J.: Deformable convnets v2: More deformable, better results. In: Proceedings of the IEEE/CVF conference on computer vision and pattern recognition. pp. 9308--9316 (2019)

\end{thebibliography}

\end{document}